\documentclass[10pt, a4paper]{article}

\usepackage{booktabs}   
\usepackage{tabularx}
\usepackage{makecell}
\usepackage{array}
\newcolumntype{C}[1]{>{\centering\arraybackslash}m{#1}}
\usepackage{subcaption}  
\usepackage{multirow}
\usepackage{amssymb}    
\usepackage{caption}
\usepackage{comment}
\usepackage{cite}
\usepackage{xcolor}
\usepackage{amsmath,amssymb,amsfonts}
\usepackage{algorithmic}
\usepackage{graphicx}
\usepackage{textcomp}
 \usepackage{booktabs}
\usepackage{tabularx}
\usepackage{xcolor}
\usepackage{threeparttable}
\usepackage{float}
\usepackage{url}
\usepackage{enumitem}  
\usepackage{xspace}  
\usepackage[final]{lrec2026} 

\newcommand{\datafull}{{\textsc Spoken DialogSum}}

\title{\datafull{}: An Emotion-Rich Conversational Dataset for Spoken Dialogue Summarization}

\name{\large\textbf{Yen-Ju Lu*, Kunxiao Gao*, Mingrui Liang*, Helin Wang, Thomas Thebaud,} \\[\smallskipamount]
\large\textbf{Laureano Moro-Velazquez, Najim Dehak, and Jesús Villalba}} 

\address{Center for Language and Speech Processing, Johns Hopkins University, \\
         Baltimore, MD, USA\\
         \{ylu125, kgao9, mliang17, hwang258, tthebau1, laureano, ndehak3, jvillal7\}@jhu.edu\\}

\abstract{
Recent audio language models can follow long conversations. However, research on emotion-aware or spoken dialogue summarization is constrained by the lack of data that links speech, summaries, and paralinguistic cues. We introduce \datafull{}, the first corpus aligning raw conversational audio with factual summaries, emotion-rich summaries, and utterance-level labels for speaker age, gender, and emotion. The dataset is built in two stages: first, an LLM rewrites DialogSum scripts with Switchboard-style fillers and back-channels, then tags each utterance with emotion, pitch, and speaking rate. Second, an expressive TTS engine synthesizes speech from the tagged scripts, aligned with paralinguistic labels. \datafull{}   comprises 13,460 emotion-diverse dialogues, each paired with both a factual and an emotion-focused summary. We release an online demo at \url{https://fatfat-emosum.github.io/EmoDialog-Sum-Audio-Samples/}, with plans to release the full dataset in the near future. Baselines show that an Audio-LLM raises emotional-summary ROUGE-L by 28\% relative to a cascaded ASR-LLM system, confirming the value of end-to-end speech modeling. \\ 
\newline  }

\begin{document}

\maketitleabstract

\renewcommand{\thefootnote}{}
\footnotetext{$^*$~denotes equal contribution.}
\renewcommand{\thefootnote}{\arabic{footnote}}

\section{Introduction}
Recent progress in Audio-LLMs—such as WavLLM~\cite{hu2024wavllm}, SALMONN~\cite{tangsalmonn}, Qwen-Audio~\cite{chu2023qwen}, and LTU-AS~\cite{gong2024listen}—demonstrates the feasibility of directly modeling speech for downstream language tasks, from translation to question answering. However, most of the existing benchmarks target a single task (e.g. ASR on LibriSpeech~\cite{panayotov2015librispeech}, emotion recognition on IEMOCAP~\cite{busso2008iemocap}). 
Even when multiple tasks are merged in a single model, these abilities are separately trained and combined with different prompts, but omit the interaction between semantic content and acoustic information.
%
Therefore, we propose \datafull{}, the first large-scale spoken dialogue summarization corpus that is paired with both text-based and emotion-rich summaries based on paralinguistic information.

 
Dialogue summarization datasets such as SAMSum~\cite{gliwa2019samsum} and DialogSum~\cite{chen2021dialogsum} drive advances in text-based summarization. However, they rely solely on transcripts of written dialogues. In contrast, spontaneous-speech corpora such as SwitchBoard~\cite{godfrey1992switchboard}, MELD~\cite{poria2019meld} capture genuine turn-taking and vocal signals but lack human-labeled summaries altogether. 
For example, DialogSum provides concise summaries of daily-life dialogues but originates from scripted transcriptions with no backchannels or disfluencies. Therefore, it fails to reflect the actual speakers' interaction. 

To address this gap, we built a framework that transforms DialogSum’s transcriptions into rich annotated speech interactions as \datafull{}. 
Inspired by the post-process in Behavior-SD \cite{lee2025behavior}, our pipeline proceeds in three steps:
First, we apply an LLM as a style-conversion model to process the dialogues with real conversational transcript examples from SwitchBoard. We rewrite each scripted dialogue to include natural disfluencies, fillers, and natural phrasing.
Next, we further insert backchannels
at contextually appropriate points in the dialogues as listener engagement.
Lastly, we assign one overall emotion style and generate an emotion-focused summary that complements the primary summary for each dialogue.
We synthesize emotion-rich, high-fidelity speech for over 13K dialogues ($\sim$165 hours) using Zonos~\cite{zonos} as the TTS model with 20K clean speech prompts annotated by age group and gender from GigaSpeech~\cite{gigaspeech}.

\begin{figure*}[t]
    \centering
    \includegraphics[width=\linewidth]{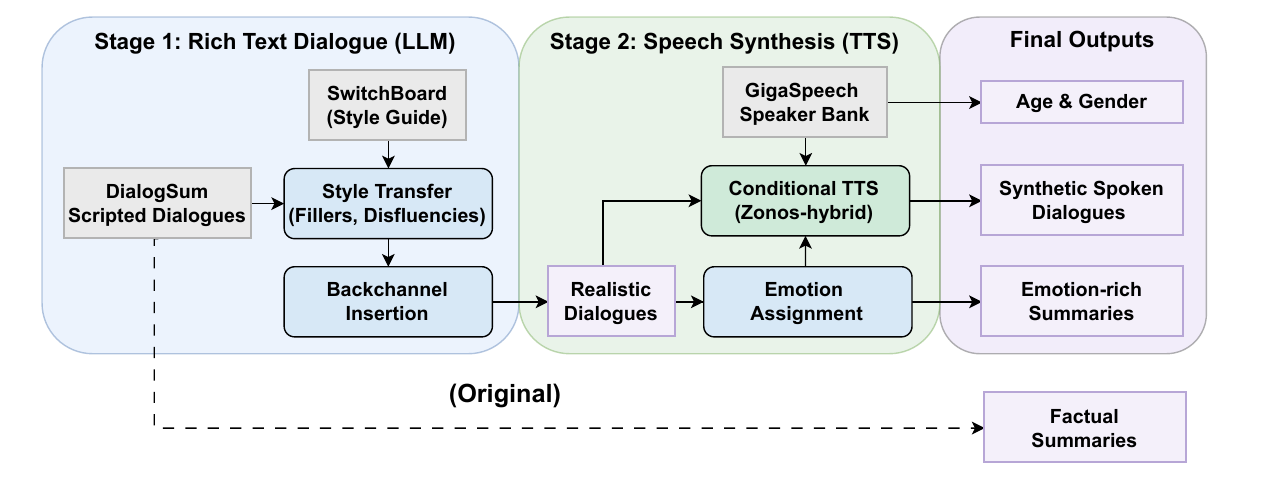}
    \vspace{-7mm}
    \caption{\datafull{} pipeline. 
    Stage~1 rewrites DialogSum scripts with Switchboard-style fillers and backchannels for realistic dialogues. Stage~2 synthesizes expressive speech with emotion and prosodic labels, producing aligned factual and emotion-rich summaries with speaker attributes.}
    \label{fig:pipeline}
    \vspace{-5mm}
\end{figure*}

\datafull{} is the first corpus to pair raw multi-speaker audio with both factual and emotion-rich summaries while also providing utterance-level labels for speaker emotion, gender, and age.  We benchmark three complementary tasks: (1) text-only factual summarization, (2) cross-modal emotion-rich summarization, and (3) acoustic-only paralinguistic-attribute classification. We evaluate two modeling paradigms: a cascaded ASR → LLM pipeline and an end-to-end Audio-LLM that consumes raw waveforms plus extracted paralinguistic cues.  
Experiments show that the Audio-LLM improves ROUGE-L on emotion-rich summarization by 29\% over the cascaded baseline, when evaluated against emotion-rich references derived from speech emotion labels. Taken together, these results demonstrate the value of joint semantic and acoustic modeling across all three tasks.


\section{Related Work}

\subsection{Text‐based Dialogue Summarization}
Existing dialogue summarization benchmarks focused on text-based summarization. The SAMSum corpus provides 16K messenger‐style dialogues with abstractive summaries, highlighting challenges such as informal language, multiple speakers, and implicit context \cite{gliwa2019samsum}. 
DialogSum is a multi-turn dataset of real-life spoken dialogues drawn from DailyDialog~\cite{li2017dailydialog}, DREAM~\cite{sun2019dream}, MuTual~\cite{cui2020mutual}, and an English-speaking practice website, covering daily-life topics such as education, work, and healthcare, with conversations between friends, colleagues, and service providers and customers~\cite{chen2021dialogsum}. 
Large‐scale benchmarks such as MediaSum (463K media‐interview transcripts) and SummScreen (TV episode transcripts) demonstrate the continued need for entity tracking and role‐bias modeling in dialogue summarizers~\cite{zhu2021mediasum,chen2022summscreen}.
To address low-resource scenarios, LLMs are further applied for data synthesis in creating new dialogues or summaries~\cite{he-etal-2024-semi,lu2025mutual}. Moreover, even without any few-shot dialogue–summary pairs, directly generating dialogues via LLMs is effective~\cite{suresh2025diasynth,lu2025paired}.
%


\begin{table*}[t]
  \centering
  \caption{System Prompts for Dialogue Processing}
  \vspace{-3mm}
  \begin{tabularx}{\textwidth}{@{} l X @{}}
    \toprule
    \textbf{Steps} & \textbf{Prompt} \\
    \midrule
    \parbox[t]{1.8cm}{Style \\ Transfer} &

You are a dialogue-style expert. Rewrite the \textbf{Original Dialogue} so it sounds like the provided \textbf{Target Style Dialogue}: preserve every speaker, line order, and meaning, while imitating the reference snippet’s use of natural fillers, mild hesitations, and brief feedback. The result should read like a smooth, casual conversation.
    \\
    \midrule
    \parbox[t]{1.8cm}{Backchannel \\ Insertion} &
    You are a back-channel expert. Insert brief, context-relevant acknowledgements into the \textbf{Original Dialogue} so it matches the spontaneous style of the provided \textbf{Reference Dialogue}. Keep every speaker line and word order unchanged; place the back-channels only at natural pauses, use them sparingly, and ensure they fit the reference tone. \\
&  \textbf{Format}: PersonX: [first part] {{PersonY: [short reaction]}} PersonX: [rest]
    \\

    \midrule
    \parbox[t]{1.8cm}{Emotion \\ Assignment} &
Analyze the dialogue’s emotions and deliver two outputs:
(1) One sentence that sums up the \textbf{Overall Emotional Tone} while mentioning each speaker’s action.
(2) \textbf{For Every Utterance}, return a JSON object exactly like:

$\{\{$"utterance": "$<$utterance$\_$text$>$", "emotion": "$<$one of 8 emotions$>$", "vector": [one-hot in [Hap, Sad, Disg, Fear, Surp, Angr, Other, Neut]], "pitch": $<0/1/2>$, "speaking rate": $<0/1/2>$ $\}\}$
Use Hap, Sad, Disg, Fear, Surp, Angr whenever possible; choose Neutral only for emotion-free statements and Other only if the utterance is nonsensical. Pitch 0/1/2 = calm / neutral / expressive; rate 0/1/2 = slow / normal / fast    \\
    \bottomrule
  \end{tabularx}
  \label{tab:long_prompts}
  \vspace{-4mm}
\end{table*}

\subsection{Spoken Dialogue Corpora with Prosodic Information}
Various speech‐based datasets support prosodic analysis. Switchboard-NXT extends the Switchboard telephone corpus with intonation labels, disfluencies, and dialogue acts for prosodic turn‐taking studies~\cite{calhoun2010nxt}. The Santa Barbara Corpus provides face-to-face dialogues annotated for pauses, emphasis, and overlap~\cite{du2000santa}. 
Traditional corpora such as the London–Lund Corpus (LLC) and IViE offer tone‐unit and prominence markings across dialects~\cite{greenbaum1990london,grabe2003ivie}. 
For summarization, AMI is a classic small-scale benchmark, containing less than 300 noisy, overlapping recordings of long-form meetings~\cite{carletta2005ami}.
%
\subsection{Conversational Dialogue Synthesis}
To make synthetic speech more natural and interactive, recent TTS and feedback‐modeling inject spontaneous phenomena and listener reactions. Style‐transfer TTS systems like AdaSpeech 3 convert reading‐style voices with filled‐pause predictors and duration experts to add rhythmic variation \cite{yan2021adaptive}. 
Backchannel models \cite{Ruede2019} and Context‐Aware Backchannel Prediction \cite{park-etal-2024-improving} predict both timing and type of listener responses. Integrated approaches further include speaker personality and topic \cite{park2024backchannel}. 
Behavior-SD~\cite{lee2025behavior} extends this direction by introducing a large-scale synthetic dialogue dataset with a wide range of spontaneous speaker behaviors and listener responses for training realistic dialogue writing models.


\section{Realistic Spoken Dialogue Data Generation}
We generate the \datafull{} dataset using a three-stage conversion: Style Transfer, Backchannel Insertion, and Emotion Assignment. Prompts are listed in Table~\ref{tab:long_prompts}.

\subsection{Rich Text Dialogue Generation}

\subsubsection{Style Transfer}
The dialogues in the DialogSum dataset are scripted and lack natural hesitation, unlike real-world conversations. To address this, we first adapt them using Switchboard-style examples, creating more realistic and interactive dialogues that still align with their original summaries.
We use a pre-trained instructed LLM model (LLAMA3.3 70B)~\cite{dubey2024llama} to conduct the style transfer. 
Using a Switchboard sample as a style guide, we prompt the LLM to insert similar fillers and hesitations, transforming the scripted lines into natural-sounding dialogue.


\subsubsection{Backchannel Insertion}
The style-transfer step ensures that the LLM generates the same number of utterances (i.e., sentences or phrases) as the original script, maintaining alignment between the transformed and source versions.
To make the conversations more interactive, we instructed the model to insert interruptions while the other speaker is talking. 
We use a special symbol $\{\text{X: backchannel}\}$ as the insertion of mid-turn back-channels as introduced in ~\cite{lee2025behavior}. To prevent the model from repeatedly using the same interruption words, we provide examples from Switchboard dialogues to guide more varied and natural backchannel selection.
Since interruptions typically occur while the other speaker is talking, we design the backchannel utterances to overlap with the speaker’s speech. This makes the dialogues more realistic and also increases the difficulty for the model to understand them.

\subsubsection{Dialogue Evaluation}

\begin{table}[h!]
\caption{Model-based evaluation results (mean) for DialogSum, Switchboard, and \datafull{}. The evaluated metrics are Nat. (Oral Naturalness), Flo. (Conversational Flow), and Coh. (Topical Coherence and Focus).}
\label{tab:spoken_results}
\vspace{-3mm}
\centering
\begin{tabular}{lcccc}
\hline
\textbf{Dialogues} & \textbf{Nat.} & \textbf{Flo.} & \textbf{Coh.} & \textbf{Avg} \\
\hline
DialogSum  & 3.86 & 4.13 & \textbf{4.59} & 4.19 \\
Switchboard      & 4.25 & 3.71 & 4.11 & 4.02 \\
\datafull{} & \textbf{4.81} & \textbf{4.15} & 4.49 & \textbf{4.48} \\
\hline
\end{tabular}
\vspace{-4mm}
\end{table}

After conducting style transfer and backchannel insertion using LLAMA3.3, we evaluated the generated dialogues with GPT-4o-mini to avoid self-bias, since LLMs often favor their own outputs~\cite{panickssery2024llm,panicksseryllm}. Using a different model reduces this effect and provides a more reliable comparison across corpora.

As shown in Table~\ref{tab:spoken_results}, \datafull{} achieves a significantly higher score in Oral Naturalness (4.81) compared with both the source DialogSum corpus (3.86) and the Switchboard reference (4.25). The Conversational Flow metric also improves to 4.15, outperforming the other two dialogue corpora. These gains can be attributed to the inclusion of natural backchannel behaviors, which make the dialogues sound more interactive and human-like.
In contrast, the Topical Coherence and Focus score slightly decreases compared to the original DialogSum from 4.59 to 4.49. This is expected since the inserted backchannels occasionally interrupt or fragment the topical continuity of an exchange, leading the model to perceive a small reduction in overall coherence despite improved conversational realism.
Overall, \datafull{} provides a highest average scores at 4.48 compare to both the original dialogue (4.19) and their target style reference (4.02).

\subsection{Spoken Dialogue Generation}
In this section, we introduce our emotion-rich, realistic spoken dialogue generation pipeline: speaker bank construction, conditional TTS synthesis with prosodic adjustments, and timing‐driven overlap placement. 
\label{sec:Spoken_Dialogue_Generation_from_Text}
\begin{table}[t]
  \centering
  \caption{Annotated variables and categories for GigaSpeech}
  \vspace{-2mm}
  \label{tab:gigaspeech_variables}
  \begin{tabular}{@{} p{0.24\linewidth} p{0.7\linewidth} @{}}
    \toprule
    \textbf{Variable}        & \textbf{Categories}                                                                                           \\
    \midrule
    \textbf{Age}             & child, teenager, young adult, middle-aged adult, elderly                                                      \\[0.5ex]
    \textbf{Gender}          & male, female, unknown                                                                                         \\[0.5ex]
    \textbf{Pitch}          & very low-pitch, low-pitch, slightly low-pitch, moderate pitch, slightly high-pitch, high-pitch, very high-pitch \\[0.5ex]
    \textbf{Expressive.}   & very monotone, monotone, slightly expressive and animated, expressive and animated, very expressive and animated   \\[0.5ex]
    \textbf{Speaking Rate}   & very slowly, slowly, slightly slowly, moderate speed, slightly fast, fast, very fast                           \\
    \bottomrule
  \end{tabular}
  \vspace{-2mm}
\end{table}

\subsubsection{Speaker bank construction} 

We annotate age, gender, pitch, expressiveness of tone, and speaking rate for GigaSpeech following \cite{wang2025capspeechenablingdownstreamapplications}. Table~\ref{tab:gigaspeech_variables} lists the categories. 
Speaker demographics are derived using a pre-trained Wav2Vec2-based age and gender estimator\footnote{\url{https://github.com/audeering/w2v2-age-gender-how-to}} \cite{burkhardt2023age}.
Following Parler-TTS \cite{DBLP:journals/corr/abs-2402-01912}, pitch and expressiveness are measured using speaker-level mean and utterance-level standard deviation of pitch, computed with PENN\footnote{\url{https://github.com/interactiveaudiolab/penn}}.
The speaker-level mean is used to generate a label for speaker pitch relative to gender, and the standard deviation is used as a proxy for how monotone or animated the utterance is.
Speaking rate is calculated by dividing the number of phonemes in the transcription by the total duration, excluding any silences.

\begin{figure}[tbp]
  \centering
  \includegraphics[width=\linewidth]{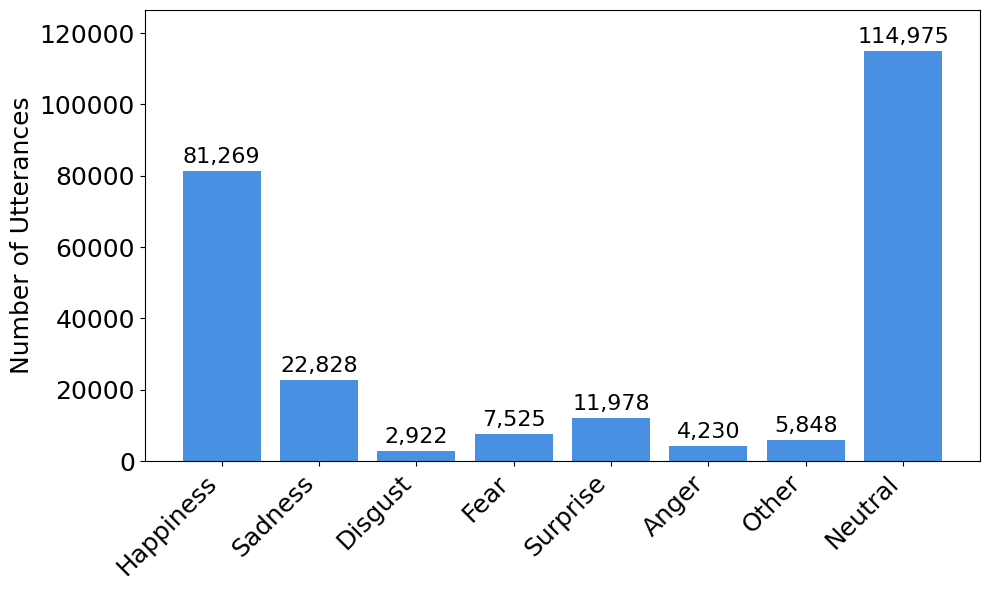}
  \vspace{-6mm}
  \caption{Distribution of utterance-level emotion labels}
  \label{fig:emotion}
  \vspace{-4mm}
\end{figure}

\subsubsection{Emotion Assignment}
To generate expressive, speaker-aware speech, we annotate each utterance with sentence-level emotion, pitch, and speaking rate. These annotations are generated using GPT-4o-mini, which is prompted with the complete dialogue along with its turn-by-turn structure. GPT is instructed to (1) produce a concise emotional summary of the dialogue and (2) assign one of eight canonical emotions (Happiness, Sadness, Disgust, Fear, Surprise, Anger, Other, or Neutral) to each utterance. These emotion labels are encoded as 8-dimensional one-hot vectors, which serve as input to the TTS model.

In addition to emotion, we extract prosodic cues from the dialogue context. Specifically, GPT is prompted to estimate pitch standard deviation and speaking rate for each utterance in the dialogue. Both are discretized into three categories—low (0), medium (1), and high (2)—to match the expected input range of the TTS model. These predictions are based on the perceived tone, formality, and engagement level of the speakers. The prompts used to derive both emotion and prosodic annotations are presented in Table~\ref{tab:long_prompts}.

The full set of style parameters (emotion vector, pitch, and speaking rate) are subsequently used as conditioning inputs to the multi-speaker TTS model described in Section~\ref{sec:Spoken_Dialogue_Generation_from_Text}, enabling generation of speech that is not only intelligible but also emotionally and prosodically appropriate.

\subsubsection{Conditional TTS model} To synthesize expressive multi-speaker dialogue audio, we adopt Zonos-hybrid, a conditional TTS model whose SSM‑Hybrid backbone interleaves Mamba‑style state‑space blocks with standard Transformer layers \cite{zonos}. 
Zonos supports speaker adaptation, enabling fine-grained control over style, emotion, and voice identity. For our experiments, we leverage this by conditioning on speakers randomly selected from a bank of $20,385$ voices derived from GigaSpeech (Table~\ref{tab:gigaspeech_variables})~\cite{gigaspeech}.

To further improve the stability, expressiveness, and quality of our synthetic speech, we carefully curate the pool of speech-prompt segments, selecting only those longer than 5 seconds and sourced from audiobook or podcast recordings. These sources typically offer lower noise levels and higher recording fidelity compared to more variable platforms like YouTube. From this filtered set, we further restrict our selection to recordings with speech monotony values classified into one of four categories (“\textit{very expressive and animated}”, “\textit{expressive and animated}”, “\textit{slightly expressive and animated}”, or “\textit{monotone}”), deliberately excluding those that are excessively flat or monotonous.

\begin{table}[tbp]
  \centering
  \caption{Overall statistics of the dataset}
  \vspace{-2mm}
  \begin{tabular}{@{} l r @{}}
    \toprule
    \multicolumn{2}{@{}l}{\textbf{Statistics}} \\
    \midrule
    \# dialogues                      & 13{,}460      \\
    \# utterances                     & 
     251.6K \\
    total duration (hr)               & 159.87        \\
    avg.\ duration per dialogue (s)              & 42.76~$\pm$~20.41    \\
    avg.\ duration per uttr.\ (s)   & 2.34~$\pm$~2.18    \\
    avg.\ words per uttr.             & 8.53~$\pm$~8.03 \\
    \# speech prompts (M/F)             & 14,742 / 12,324 \\
    \bottomrule
  \end{tabular}
  \label{tab:dataset-stats}
  \vspace{-4mm}
\end{table}

Once suitable prompts have been assigned in given dialogues, we inject the previously generated emotion vectors into Zonos-hybrid. To compensate for the TTS model’s tendency toward under-expressive affect in short utterances, we deliberately elevate the baseline inputs for pitch standard deviation and speaking rate. Concretely, we map “low”, “medium”, and “high” pitch levels to 60.0, 85.0, and 110.0, respectively, and analogous speaking-rate levels to 15.0, 18.0, and 21.0 (in units of phonemes per second). Additionally, we observed that Zonos can truncate ultra-short backchannel phrases (e.g., “got you”) too abruptly; to mitigate this, every backchannel utterance is synthesized at the lowest speaking-rate (0), and we append one second of silence after these extremely brief segments. By carefully filtering reference prompts, adjusting prosodic inputs, and introducing silence padding, we achieve more natural, emotionally resonant, and smoothly transitioned multi-speaker dialogue synthesis.

\begin{table*}[htbp]
  \centering
  \caption{
  Statistics (a) and human evaluation results (b) of spoken dialogue datasets.}
  \label{tab:human_overall}
  
  \begin{subtable}[t]{\textwidth}
    \centering
    \caption{Dataset statistics include full-duplex support, behavior labels, public availability, recording type, number of dialogues, and total audio duration (hours).}
    \vspace{-2mm}
    \label{tab:human_data_stats_sub}

    \setlength{\tabcolsep}{5pt}

    {\small
    \begin{tabularx}{\textwidth}{@{} 
        l
        c   
        c   
        c   
        c   
        c   
        c   
        c   
      @{}}
      \toprule
      \textbf{Dataset}
        & \textbf{Full‐Duplex}
        &     \makecell[c]{\textbf{Emotion }\\\textbf{Label}}
        & \makecell[c]{\textbf{Summ. +}\\\textbf{Emo.Summ.}}
        & \makecell[c]{\textbf{Public}\\\textbf{Access}}
        & \textbf{Category}
        & \textbf{\# Dialogues}
        &     \makecell[c]{\textbf{Audio }\\\textbf{(hrs)}}\\
      \midrule
      Switchboard
        & 
        & 
        & 
        & 
        & recorded
        & 2{,}400
        & 260 \\
      MELD
        & \checkmark
        & \checkmark
        & 
        & \checkmark
        & recorded
        & 1{,}400
        & 12 \\
      DailyTalk
        & \checkmark
        & \checkmark
        & 
        & \checkmark
        & recorded
        & 2{,}541
        & 20 \\
      Behavior‐SD
        & \checkmark
        & 
        & 
        & \checkmark
        & TTS‐converted
        & 108{,}174
        & 2{,}164 \\
      \midrule
      Spoken DialogSum
        & \checkmark
        & \checkmark
        & \checkmark
        & \checkmark
        & \textbf{TTS‐converted}
        & \textbf{13{,}640}
        & \textbf{160} \\
      \bottomrule
    \end{tabularx}
    }

    \vspace{0.5ex}
    \begin{minipage}{\textwidth}
    \end{minipage}

  \end{subtable}

  \vspace{1em}  

  \begin{subtable}[t]{\textwidth}
  \centering
  \caption{Human evaluation metrics report average scores for naturalness, emotion expressivity, emotion consistency, sound quality, and the overall average.}
  \vspace{-2mm}
  \label{tab:human_scores_sub}
    \begin{tabular}{@{} l c c c c c @{}}
      \toprule
      \textbf{Dataset}
        & \textbf{Naturalness}
        & \textbf{Emo. Expr.}
        & \textbf{Emo. Cons.}
        & \textbf{Sound Quality}
        & \textbf{Avg.} \\
      \midrule
      Switchboard
        & 3.61
        & 3.53
        & 3.76
        & 2.88
        & 3.45 \\
      MELD
        & \textbf{4.06}
        & \textbf{4.46}
        & \textbf{4.36}
        & 3.58
        & \textbf{4.12} \\
      DailyTalk
        & 2.70
        & 3.28
        & 3.36
        & \textbf{4.73}
        & 3.52 \\ \hline
      Behavior‐SD
        & 2.84
        & 2.83
        & 2.97
        & \textbf{4.60}
        & 3.31 \\
      \datafull{}
        & \textbf{3.64}
        & \textbf{3.84}
        & \textbf{3.75}
        & 3.89
        & \textbf{3.78} \\
      \bottomrule
    \end{tabular}
  \vspace{-4mm}
\end{subtable}

\end{table*}
\subsubsection{Timing-driven utterance placement} When merging interrupt and backchannel segments into the original audio, we adjust their timing to mirror natural conversations. To guide placement, we use timing statistics from the real‐world spoken dialogue corpus CANDOR~\cite{CANDOR}. This corpus shows that interruptions typically occur in a normal distribution  $N\bigl(0.45\,\mathrm{s},\,0.05\,\mathrm{s}\bigr)$ before the previous speaker finishes~\cite{CANDOR}. To account for the typical lead-in and trailing silences of an utterance, and to create a more perceptible overlap, we insert an additional 1-second buffer in interruptions, placed 1.5 seconds before the end of the host’s turn. For backchannels, the delay is drawn from a normal distribution $N\bigl(0.2\,\mathrm{s},\,0.02\,\mathrm{s}\bigr)$ after the previous speaker's turn \cite{CANDOR}. Because utterances naturally include brief leading and trailing silences, we treat those silences as natural delay and place the backchannel at the start of the following speaker’s turn. This combination of statistical timing and silent padding better replicates the flow of spontaneous dialogue.

\section{The Spoken DialogSum Datasets}

\datafull{} comprises 13,460 multi‐speaker dialogues and 251,575 utterances, totaling roughly 160 hours of audio. Each dialogue is accompanied by both a concise summary and an emotion‐rich summary. The details of statistics are shown in Table~\ref{tab:dataset-stats}. The 160 hours of well‐curated, speech‐style–annotated audio is one of the largest emotion‐rich, full‐duplex spoken dialogue datasets with summaries available. Figure~\ref{fig:emotion} illustrates the utterance‐level emotion distribution: 32.3$\%$ of turns are labeled Happiness, 9.07$\%$ Sadness, 1.16$\%$ Disgust, 2.99$\%$ Fear, 4.76$\%$ Surprise, 1.68$\%$ Anger, 2.32$\%$ Other, and 45.72$\%$ Neutral. Unlike many existing corpora that skew heavily toward Neutral or lack fine‐grained affect, \datafull{} shows a more balanced spread: over  40$\%$ of utterances convey clear positive (Happiness) or negative (Sadness) and about 13$\%$ of utterances convey nuanced (Surprise, Fear, and etc.) states, making it suitable for training and evaluating emotion‐aware models. Example generated dialogue audios are available\footnote{\url{https://fatfat-emosum.github.io/EmoDialog-Sum-Audio-Samples/}}.

In Table~\ref{tab:human_overall}, we report summary statistics alongside human evaluation outcomes for several spoken‐dialogue collections. Specifically, we benchmark \datafull{} against human‐recorded corpora such as Switchboard \cite{switchboard1992} and MELD \cite{poria2019meld}, human‐read conversations from DailyTalk \cite{dailytalk2023}, and synthetic dialogues from Behavior-SD \cite{lee2025behavior}. 
To gather perceptual judgments, we recruited 12 university-affiliated student raters. They rated 480 audio segments (each 20–30 seconds long) on a 1–5 scale across four criteria: Naturalness, Emotion Expressivity, Emotion Consistency, and Sound Quality. Naturalness assesses how closely prosody and pacing mimic spontaneous human speech without obvious synthesis artifacts; Emotion Expressivity determines whether the delivery is monotone or richly expressive; Emotion Consistency judges whether the emotional tone matches the content and context of the dialogue; and Sound Quality measures the degree to which recordings are free of noise and distortion and meet professional audio standards.

As shown in Table~\ref{tab:human_overall}, \datafull{} is the first large-scale spoken dialogue corpus to include emotion-rich summaries, setting it apart from existing datasets.
While MELD draws from the Friends TV show and offers highly natural, emotionally rich speech, with fine-grained emotion annotations and occasional background noise, it is limited to 12 hours of audio without available summaires.

In contrast, \datafull{} demonstrates consistently strong performance across all human evaluation criteria, with an overall average of 3.78, second only to MELD (4.12). Notably, \datafull{} achieves high ratings for naturalness (3.64) and emotion-related metrics (3.84 for expressivity, 3.75 for consistency), clearly surpassing other TTS-generated corpora such as Behavior-SD and rivaling human-read collections like DailyTalk. Furthermore, \datafull{}’s sound quality (3.89) exceeds that of large recorded dialogue corpora like Switchboard (2.88) and MELD (3.58), highlighting its robustness despite being synthesized.

Beyond perceptual strengths, \datafull{} offers approximately 160$\mathrm{h}$ of audio, far exceeding MELD’s 12$\mathrm{h}$ and DailyTalk’s 20$\mathrm{h}$, and uniquely provides per-utterance pitch-std and speaking-rate labels. By combining large scale, strong perceptual quality, rich style annotations, and dedicated emotion-focused summaries, \datafull{} is  well-suited for emotion summarization and related large-scale spoken dialogue tasks.

\begin{table*}[t]
  \centering
  \caption{Evaluation metrics and task prompts. Prompts are abbreviated here for space, full versions in Appendix.}
  \vspace{-2mm}
  \label{tab:eval-prompts}
  \begin{tabularx}{\textwidth}{@{} 
      l   
      l   
      l   
      X          
    @{}}
    \toprule
    \textbf{Category} & \textbf{Task} & \textbf{Eval.} & \textbf{Prompt (abbrev.)} \\
    \midrule
    
    \textbf{Semantic}
      & 
      \makecell[l]{Factual \\ Summarization}
      & \makecell[l]{ROUGE, \\ BERTScore}
      & ``Write a concise summary of the dialogue.'' \\
    
    \midrule
    \multirow{5}{*}{\textbf{Paralinguistic}} 
      & 
      \multirow{2}{*}{Age Prediction} 
      & \multirow{2}{*}{Acc., F1} 
      & ``Identify speaker age group (teenager / young adult / middle-aged / elderly).'' \\
    & Gender Prediction 
      & Acc., F1 
      & ``Identify speaker gender (male / female).'' \\
    & 
    \multirow{2}{*}{Emotion Classification} 
      & \multirow{2}{*}{Acc., F1} 
      & ``Classify conversation-level emotion (positive / negative / neutral / others).'' \\
    
    \midrule
    \makecell[l]{\textbf{Semantic $\times$} \\ \textbf{Paralinguistic}} 
      & \makecell[l]{Emotion-Rich \\ Summarization}
      & \makecell[l]{ROUGE, \\ BERTScore}
      & \makecell[l]{``Generate an emotional summary of each \\ speaker
throughout the conversation.''} \\
    
    \bottomrule
  \end{tabularx}
  \vspace{-4mm}
\end{table*}
\section{Experimental Setup}
As shown in Table~\ref{tab:eval-prompts}, \datafull{} provides a three-way examination of dialogue understanding:

\noindent\textbf{Task 1 – Factual Summarization (purely semantic).} The model condenses a dialogue’s propositional content using only textual cues, evaluating its ability to perform semantic abstraction.

\noindent\textbf{Task 2 – Emotion/Gender/Age Classification (purely paralinguistic):} With transcripts removed, the model infers speaker emotion, gender, and age directly from vocal characteristics, assessing competence on paralinguistic cues alone.

\noindent\textbf{Task 3 – Emotion-Rich Summarization (semantic $\times$ paralinguistic).} 
The system must fuse lexical meaning with vocal affect, capturing \emph{what} was said and \emph{how} it was expressed, so that the summary reflects both semantic content and emotional nuance, thereby testing cross-modal integration.

Together, tasks 1–3 form a continuum from text-only reasoning to multimodal fusion and audio-only interpretation, giving \datafull{} a broad view of multimodal dialogue comprehension.

\subsection{Baseline Models}
\noindent\textbf{LLM (Transcript-Only).}
We bypass audio entirely and feed the reference transcripts to \textsc{LLaMA-2-7B-chat}~\cite{genai2023llama}.  
The model then produces both factual and emotion-aware summaries.


\noindent\textbf{Whisper + LLM (Cascaded).} 
Whisper Large V2 first transcribes the speech, and \textsc{LLaMA-2-7B-chat} summarizes the resulting text.  
This pipeline lets us separate ASR quality from downstream language understanding.


\noindent\textbf{WavLLM (End-to-End).} 
The architecture consists of a Conformer encoder that extracts acoustic features, which are then fused into a \textsc{LLaMA} decoder through dual cross-attention blocks. This design forms a fully speech-to-text framework.

\noindent\textbf{Qwen-Audio-Chat (End-to-End).} 
The model consists of a Whisper encoder that provides latent speech representations to a Qwen language model through a lightweight fusion adapter, enabling integration of acoustic and semantic information.  

\noindent\textbf{Audio-Flamingo3 (End-to-End).}
Built on AF-Whisper, it jointly encodes speech, sound, and music, projecting them through adaptor layers into a Qwen-2.5-7B decoder to achieve seamless cross-modal reasoning.

\noindent\textbf{LTU-AS (End-to-End).} 
Speech is processed by a frozen Whisper encoder and passed into a \textsc{LLaMA} decoder through a time- and layer-wise Transformer bridge. This keeps the ASR front end fixed while introducing alignment layers for modality fusion.

\noindent\textbf{SALMONN (End-to-End).}
The architecture combines a frozen Whisper encoder with a Vicuna decoder, linked by a \textit{Q-Former} alignment module. This configuration preserves strong language priors while establishing an audio–text interface.

\noindent\textbf{{Wav2Vec2-Based}}.
We use a wav2vec 2.0–based model, fine‐tuned on aGender~\cite{burkhardt2010database}, Mozilla Common Voice~
\cite{ardila2020common}, TIMIT~\cite{garofolo1993timit}, and VoxCeleb 2~\cite{chung2018voxceleb2}, to perform age and gender classification tasks.

\setlength{\tabcolsep}{10pt}
\begin{table*}[htbp]
\centering
\small
\caption{Performance on the dialogue summarization and emotion-rich summarization tasks. All scores are shown as percentages. Bold indicates the best result; underline indicates second-best.}
\vspace{-2mm}
\label{tab:summary_comparison}
\begin{tabular}{@{}lcccccccc@{}}
\toprule
\textbf{Model} & \multicolumn{4}{c}{\textbf{Dialogue Summarization}} & \multicolumn{4}{c}{\textbf{Emotion-Rich Summarization}} \\
\cmidrule(rl){2-5}\cmidrule(l){6-9}
& R-1${\uparrow}$ & R-2${\uparrow}$ & R-L${\uparrow}$ & F1${\uparrow}$
& R-1${\uparrow}$ & R-2${\uparrow}$ & R-L${\uparrow}$ & F1${\uparrow}$ \\
\midrule
Transcription + LLaMA 2 & \underline{28.0} & \textbf{10.1} & \underline{21.8} & \textbf{87.6}
                    & 25.2 & 1.1 & 23.1 & 88.5 \\
Whisper + LLaMA 2     & \textbf{28.6} & \underline{9.8} & \textbf{22.0} & \underline{87.0}
                    & 24.4 & 0.8 & 21.6 & 88.0 \\
\midrule
WavLLM              & 27.9 & 8.5 & 21.5 & 86.9
                    & \underline{33.4} & \underline{8.8} & \underline{27.8} & \underline{91.1} \\
Qwen-Audio                & 22.2 & 6.6 & 17.1 & 85.7
                    & 18.5 & 1.4 & 15.9 & 87.2 \\
Audio-flamingo3     & 26.9 & 7.3 & 21.1 & 86.9
                    & 22.4 & 4.4 & 18.2 & 88.8 \\
LTU-AS              & 20.5 & 5.4 & 15.4 & 85.5
                    & 18.1 & 1.2 & 15.4 & 86.8 \\
SALMONN-7B          & 17.6 & 5.4 & 13.5 & 85.0
                    & 19.9 & 5.2 & 17.1 & 87.5 \\
SALMONN-13B         & 22.7 & 6.7 & 17.8 & 86.4
                    & \textbf{35.9} & \textbf{13.3} & \textbf{30.8} & \textbf{91.5} \\
\bottomrule
\end{tabular}
\vspace{-4mm}
\end{table*}
\setlength{\tabcolsep}{6pt}

\subsection{Evaluation Framework}
We perform our evaluation on the \datafull{} test split, which comprises 500 dialogues, each paired with three human-written summaries. The dialogue summarization score is computed by averaging the results across those three reference summaries. Table~\ref{tab:eval-prompts} shows the abbreviated prompts used in evaluation.

\noindent\textbf{Dialogue Summarization.} We evaluate whether the systems can generate concise and coherent summaries based on their semantic content. For text-only models, the input is the ground truth transcript, while for all other models, the full dialogue audio is provided. All models are prompted with the same instruction, and are expected to produce a 2--3 sentence summary. To assess summary quality, we use ROUGE-1, ROUGE-2, ROUGE-L, and BERTScore. 
Each generated summary is compared against three ground truth references, and the final score is computed by their average.

\noindent\textbf{Emotion-Rich Dialogue Summarization.} 
To test the model's performance on emotional reasoning, we give the full spoken dialogue as input, and the model is prompted to generate a one-sentence summary describing the emotional expression of each speaker. To assess whether the model can reliably capture such speaker-level affective cues, we use the same automatic metrics as in dialogue summarization—ROUGE-1, ROUGE-2, ROUGE-L, and BERTScore against the corresponding emotion-rich summary.


\noindent\textbf{Paralinguistic Attribute Prediction.} 
This task is designed to assess whether the models are able to evaluate acoustic cues for identifying speaker-level attributes—specifically, \textit{age group}, \textit{gender}, and \textit{emotion}—from full spoken dialogues. 
Since these attributes rely heavily on prosodic and acoustic features that are absent in pure text, we exclude the text-only and cascaded models from this evaluation. Each dialogue is fed into the model as a whole, and the model is prompted to predict the \textit{age} and \textit{gender} of both speakers, as well as the overall \textit{emotion} expressed in the conversation. The age group is selected from four categories: \textit{teenager}, \textit{young adult}, \textit{middle-aged adult}, and \textit{elderly}; gender is classified as either \textit{male} or \textit{female}; and emotion is predicted as one of \textit{positive}, \textit{negative}, or \textit{neutral}. For evaluation, we report both accuracy and weighted F1-score to reflect robustness and account for class imbalance.

\section{Results}

\subsection{Dialogue and Emotion-Rich Summarization Results}
Table~\ref{tab:summary_comparison} compares the baseline models on evaluation axes involving semantic reasoning, both in isolation and when combined with paralinguistic cues. For Task 1 (purely semantic reasoning), where only the semantic content matters, the transcript-only LLaMA-2 and its cascaded Whisper + LLaMA-2 variant top the leaderboard, confirming that text-centric LLMs are most effective when no paralinguistic cues are required.
When we switch to Task 3 (semantic $\times$ paralinguistic interaction)—emotion-rich summarization—the ranking reverses. The audio-conditioned SALMONN-13B delivers the best overall scores, with WavLLM close behind, demonstrating their ability to fuse acoustic affect with lexical meaning. Text-only baselines slump sharply, while cross-modal models such as Qwen-Audio, LTU-AS, and SALMONN-7B exhibit mixed gains, underlining that both architecture and training strategy influence how well semantic and acoustic evidence are integrated.
Taken together, these results validate \datafull{}’s design: Task 1 isolates a model’s semantic abstraction ability, whereas Task 3 probes its competence at weaving affective acoustics into coherent summaries.

\begin{table}[t]
\centering
\small
\caption{Accuracy(\%) and F1 (\%) on speaker-level attribute prediction tasks.}
\label{tab:attribute_prediction}
\begin{tabular}{@{}lcccc@{}}
\toprule
\multirow{2}{*}{\textbf{Dataset/Model}} & \multicolumn{2}{c}{\textbf{Age}} & \multicolumn{2}{c}{\textbf{Gender}} \\
\cmidrule(rl){2-3}\cmidrule(l){4-5}
 & Acc.${\uparrow}$ & F1${\uparrow}$ & Acc.${\uparrow}$ & F1${\uparrow}$ \\
\midrule
EMODB (Wav2Vec2) & 67.7 & 80.7 & 95.7 & 95.7 \\  \midrule
Wav2Vec2 & \textbf{66.3} & \textbf{65.2} & \textbf{95.4} & \textbf{95.4} \\  
WavLLM & 31.4 & 29.0 & {59.7} & {59.1} \\
Qwen-Audio & {48.8} & {45.0} & 51.0 & 34.5 \\
\bottomrule
\end{tabular}
\vspace{-1mm}
\end{table}
\subsection{Paralinguistic Attribute Prediction}
Task 2 evaluates a model’s ability to infer nonverbal speaker attributes including \textit{age group}, \textit{gender}, and \textit{emotion} from acoustic signals. Table~\ref{tab:attribute_prediction} shows results for age and gender classification. {Wav2Vec2} achieves the strongest performance (66.3 Acc, 65.2 F1 for age; 95.4 Acc/F1 for gender), closely matching the accuracy reported on real annotated data such as EMODB (67.7 Acc, 80.7 F1 for age; 95.7 Acc/F1 for gender). This alignment suggests that our dataset effectively reflects authentic age and gender patterns. By contrast, {WavLLM} and {Qwen-Audio} show weaker results, indicating the difficulty of capturing fine-grained speaker traits without explicit supervision.
Table~\ref{tab:emotion_4emo_comparison} presents emotion recognition in a 4-class setup. {LTU-AS} slightly outperforms {WavLLM} (49.1 Acc vs. 42.5 on IEMOCAP; 47.8 vs. 45.8 on EmoSum), and both models show consistent trends with human-labeled benchmarks, confirming that the data also captures realistic emotional cues.
Overall, Task 2 highlights that paralinguistic understanding requires more than text alone and depends on robust acoustic modeling. The close correspondence between model performance on our benchmark and real annotated corpora further validates that the dataset captures genuine speaker characteristics. 
%
%

\begin{table}[t]
\centering
\small
\caption{Comparison of Accuracy (\%) and Weighted F1 (\%) on two emotion datasets (4-emotion setup).}
\label{tab:emotion_4emo_comparison}
\begin{tabular}{@{}lcccc@{}}
\toprule
\multirow{2}{*}{Model} & \multicolumn{2}{c}{IEMOCAP} & \multicolumn{2}{c}{EmoSum} \\
\cmidrule(lr){2-3} \cmidrule(lr){4-5}
 & Acc.${\uparrow}$ & F1-W.${\uparrow}$ & Acc.${\uparrow}$ & F1-W.${\uparrow}$ \\
\toprule
WavLLM & 42.52 & 35.81 & 45.78 & 44.20 \\
LTU-AS & 49.12 & 38.45 & 47.75 & 47.65 \\
\bottomrule
\end{tabular}
\end{table}














\section{Conclusion}
We introduced \datafull{},  a large-scale benchmark that probes dialogue understanding along three separate axes:
(i) factual summarization from text only,  (ii) emotion-rich summarization that fuses lexical and acoustic cues, and (iii) acoustic-only prediction of speaker emotion, gender, and age.
To build the corpus, we transform DialogSum scripts into Switchboard-style conversations, inserting realistic back-channels and synthesizing expressive audio with a conditional TTS pipeline. We created 13,460 dialogues ($\sim$165 h) that capture authentic turn-taking, disfluencies, and emotional nuance.
Baseline experiments reveal substantial performance gaps across modeling paradigms: raw speech input with Audio-LLMs improves ROUGE-L for emotional summaries by 28\% compared to a cascaded ASR+LLM pipeline, and Wav2Vec 2.0–based classifier shows strong gains in age and gender prediction at the utterance level. 
Human evaluations further confirm that \datafull{} achieves higher naturalness and emotion consistency than prior synthetic dialogue corpora.

\section*{Ethics Statement}
Spoken DialogSum was constructed based on existing open datasets. The dialogue texts originate from publicly available corpora such as DialogSum, while the speech component is synthesized using a conditional TTS model conditioned on speaker samples from GigaSpeech. All sources are released under research licenses, and no private or personally identifiable data are included. The dataset is intended solely for academic research in speech and language processing. We caution against potential misuse, such as applying paralinguistic classifiers for demographic profiling or surveillance, which could raise ethical concerns. Our release will emphasize appropriate use for scientific purposes and transparency in data generation.

\nocite{*}

\bibliographystyle{lrec2026-natbib}
\bibliography{lrec2026-example}

\bibliographystylelanguageresource{lrec2026-natbib}
\bibliographylanguageresource{languageresource}

\end{document}